\documentclass{article}

\usepackage{arxiv}

\usepackage[utf8]{inputenc} 
\usepackage[T1]{fontenc}    
\usepackage{hyperref}       
\usepackage{url}            
\usepackage{booktabs}       
\usepackage{amsfonts}       
\usepackage{nicefrac}       
\usepackage{microtype}      
\usepackage{lipsum}
\usepackage[pdftex]{graphicx}    
\usepackage{graphicx}
\usepackage{amsmath,amssymb} 
\usepackage{color}
\usepackage{multirow}
\usepackage{makecell}
\usepackage{multirow}
\usepackage[table,xcdraw]{xcolor}
\title{Improving Unsupervised Anomaly Localization by Applying Multi-scale memories to Autoencoders}

\author{
  Yifei Yang \\
  Department of Mathematics\\
  Shandong University\\
  \texttt{yangyfaker@gmail.com} \\
   \And
 Shibing Xiang \\
  Department of Mathematics\\
  Shandong University\\
  \texttt{chibing.xiang@gmail.com} \\
  \And
 Ruixiang Zhang \\
  Zhejiang University\\
  \texttt{zrxisgood@gmail.com}
}

\begin{document}
\maketitle

\begin{abstract}
Autoencoder and its variants have been widely applicated in anomaly detection.The previous work memory-augmented deep autoencoder proposed memorizing normality to detect anomaly, however it neglects the feature discrepancy between different resolution scales, therefore we introduce multi-scale memories to record scale-specific features and multi-scale attention fuser between the encoding and decoding module of the autoencoder for anomaly detection, namely MMAE.MMAE updates slots at corresponding resolution scale as prototype features during unsupervised learning. For anomaly detection, we accomplish anomaly removal by replacing the original encoded image features at each scale with most relevant prototype features,and fuse these features before feeding to the decoding module to reconstruct image. Experimental results on various datasets testify that our MMAE successfully removes anomalies at different scales and performs favorably on several datasets compared to similar reconstruction-based methods.
\end{abstract}


\section{Introduction}
Anomaly detection is a significant problem and well-studied with numerous applications in multiple fields, such as defects detection of workpiece\cite{bergmann2019uninformed},object detection, lesion localization of medical images\cite{schlegl2017unsupervised,chen2018unsupervised,zhou2019sparse}, and also in the area of credit fraud detection\cite{zenati2018adversarially,akcay2018ganomaly}, or X-ray screening for security\cite{akcay2018ganomaly}, network intrusion detection\cite{taha2019anomaly,zenati2018adversarially}. Among these fields,Unsupervised classification and segmentation of anomalous images(or video frames) is an important and challenging task in many areas of computer vision. In automated industrial inspection scenarios, we can usually get a lot of normal images easily, but the anomalous images that can be acquired are very scarce, so we want to construct a model like autoencoder that are only trained on defect-free images.

\begin{figure}[ht]
	\centering
	 \begin{minipage}{1.\textwidth}
     \centering
	\vspace{-0.5cm}
	\includegraphics[width=0.7\columnwidth]{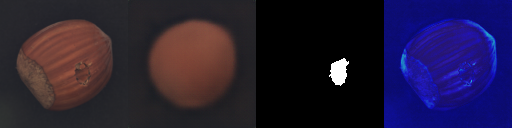} 
	\includegraphics[width=0.7\columnwidth]{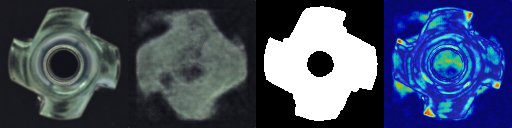}
\caption{The examples of failure reconstruction of Vanilla VAE. Columns from left to right:[1] Origin image, [2] VAE reconstruction, [3] Ground truth mask, [4] Input image with anomaly map overlay.}
\label{fig:img_reconstruct}
	\vspace{-0.8cm}
	\hspace{1.8cm}
	\end{minipage}
\end{figure}

Autoencoder (AE)\cite{goodfellow2016deep}are commonly used as a base architecture in unsupervised anomaly detection settings. They attempt to reconstruct defect-free training samples through a bottleneck (latent space).During testing, they fail to reproduce images that differ from the data that was observed during training. Anomalies are detected by a per-pixel comparison of the input with its reconstruction.Methods like this are called reconstruction-based anomaly detection.The reconstruction-based anomaly detection should follow two assumptions--One is that the model can reconstruct the normal areas of the image same as the original one,the other is that the model cant reconstruct the abnormal patterns on the original image.MemAE\cite{gong2019memorizing} and other papers\cite{zong2018deep} have found that sometimes AE can still reconstruct the abnormal inputs well, which violates the second assumption.MemAE believes that because some anomalies share common local patterns with the normal training data,or the decoder is “too strong” for decoding some abnormal encodings well.Therefore, memAE proposes a training method that induces a memory to store prototypical normal patterns of the normal training data during the training process. For test images, memAE does not feed images' encodings directly into the decoder, but instead uses it as a query to retrieve the most relevant items in the memory. These items are then aggregated and passed to the decoder, a process that removes the anomaly information from the encodings, thus preventing an strong decoder from reconstructing the anomaly.

\begin{figure*}[h]
 \centering
 \begin{minipage}{1.\textwidth}
 \centering
 \includegraphics[width=0.7\columnwidth]{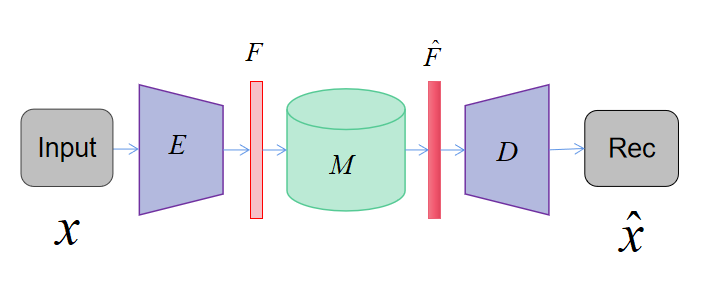}
 	\caption{The architecture of the MemAE.Given an input x,the encoding was obtained by encoder,then the anomaly removal will be performed by memory addressing.Finally, the decoder get the anomaly-free encoding and produce the de-anomalous image}
 \label{fig:memae}
 \end{minipage}
\end{figure*}

\begin{figure}[]
	\centering
	\vspace{-0.8cm}
	\includegraphics[width=120mm]{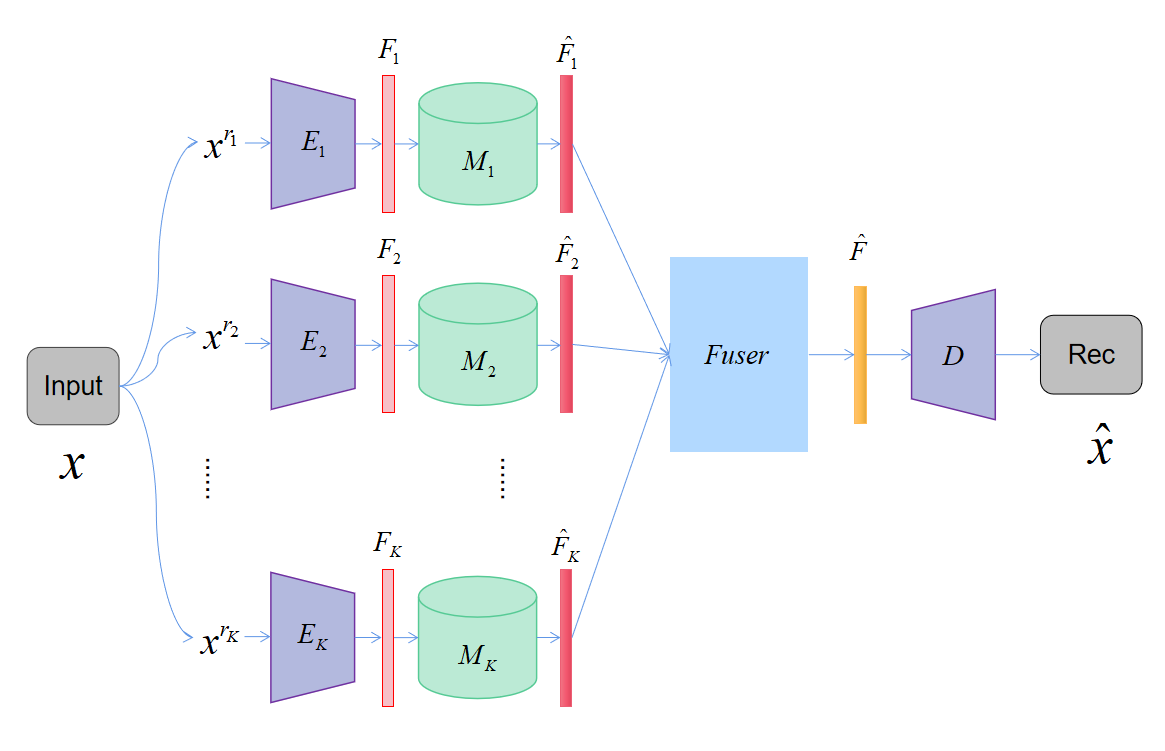} 
	\setlength{\abovecaptionskip}{-0.1cm}
	\caption{The figure shows the coarse architecture of MMAE.Taking image $x$ as input, we first resize it to resolution $r_1,r_2,...,r_K$.For each $x^{r_i}$,we get its encoding from the corresponding encoder.The encoding's potential anomalous information will then be removed through the memory addressing process, and finally the features at different levels are fused by multi-scale attention fuser to get a new feature, which is then feed into the decoder to obtain the final de-anomalous image the same resolution with input $x$.}
	\vspace{-0.1cm}
	\label{fig:architecture}
\end{figure}

However, the problem with memAE is that by replacing the original latent code with a linear combination of the fixed slots in the memory, the latent space in the memAE is a linear subspace of the AE's latent space, so that the representation ability of the model is greatly weakened. Normal images still vary significantly at different scales, but memAE, with only a limited combination of slots, sometimes find it difficult to accurately restore normal regions(see Fig.~\ref{fig:img_reconstruct}), which violates the first assumption.To mitigate this problem, we propose to use multiple memories (see Fig.~\ref{fig:architecture}), each of which stores normal data features at the corresponding scale.For each test image,our MMAE can replace the features of each scale with anomaly-free features, and fuse these features based on multi-scale attention,finally reconstruct the image using the fused feature. This strategy of separately memorizing by scale allows us to obtain a larger anomaly-free latent code space, thus satisfying both of the above assumptions.

\section{Related Works}
\subsection{Reconstruction-based anomaly Detection.}The reconstruction-based approach is applied on the assumption that anomalies cannot be accurately represented and reconstructed by models that are learned only from normal data,thus an anomaly segmentation can be obtained by a per-pixel comparison of the reconstructed image with the original input. Beyond the variants of AE like SSIM-AE\cite{bergmann2018improving},VAE\cite{kingma2013auto}, there has been some work\cite{schlegl2017unsupervised} proposing to model the manifold of the training data by a generative adversarial network(GAN)\cite{goodfellow2014generative} that is trained solely on defect-free images.The generator is able to produce realistically looking images that fool a simultaneously trained discriminator network in an adversarial way. For anomaly detection, the algorithm searches for a latent sample that reproduces a given input image. Such jobs include GANomaly\cite{akcay2018ganomaly},Sparse-GAN\cite{zhou2019sparse} and so on.

\subsection{Memory-augmented networks.}Memory network is a scheme to augment neural networks using external memory.It has attracted increasing interest for solving problems like question-answering systems \cite{sukhbaatar2015end,kumar2016ask}, summarization \cite{kim2018abstractive},image generation\cite{kim2018memorization}, and anomaly detection\cite{park2020learning}.Since this scheme can remember selected critical information, it is effective for one-shot or few-shot learning. 

\subsection{Attention mechanism}
The attention mechanism, first proposed in computer vision, allows the network to weigh information from different regions, channels, scales, and groups.In neural networks, Mnih et al.\cite{mnih2014recurrent} used an attentional mechanism on a recurrent neural network model to perform image classification. xu et al.\cite{xu2015show} uses an attentional mechanism for image content description. Wang et al.\cite{wang2019musical} utilize the attention mechanism for image inpainting.Zhang et al.\cite{zhang2020resnest} introduces split-attention in the object detection backbone.


\section{Multi-Scale Memories and Attention Autoencoder}
\subsection{Overview of the Network Architecture}

Our MMAE was derived from memAE.The memAE follows three steps when performing image reconstruction(see Fig.~\ref{fig:memae}).First,given an input $ x $, the encoder obtains the encoding of the input.Second, by using the encoded representation as a query, the memory module will remove the anomaly infomation from the encoding by retrieving the most relevant items in the memory via the attention-based addressing operator. Finally, the reconstructed anomaly-free image will be obtained by feeding the decoder with the de-anomalous encoding.

The distinction between our method and memAE is that: 1. we use multiple encoders to obtain features of the image at different scales and simultaneously utilize multi-scale memories to perform de-anomalization of features at different scales. 2. We use a multi-scale attention-based fuser to implement features fusion. The encoder and decoder will be formly described in section3.2, the memory module will be represented in section3.3 and the multi-scale fuser will be discussed detailedly in section3.4.
\subsection{Encoder and Decoder}
The encoding module is used to represent the input in an informative latent domain.We separate the latent domain into K sub-domains corresponding to K scales. The $ i $-level encoded representation performs as a query to retrieve the relevant items in the memory  $M_i$ . In our model, the encoding module can be seen as a multi-query generator. The decoder is trained to reconstruct the samples by taking the fused features as input.
Specifically speaking, we parallel K similar pipelines to form a encoding module, each pipeline takes $x_{r_i},i=1,...K$ as inputs, which is obtained by resize $x$  to resolution $r_i \times r_i$. We select a downsample factor $ \gamma \in (0,1) $ , and denote the resolution of $ x $ to be $ r = (H,W)$, then $ r_i = (H*\gamma^{i-1},W*\gamma^{i-1}) $.Then the features of the image $ x $ at different resolutions are then outputed by the corresponding pipeline.We will choose the pipeline considering the tasks we are handling, which will be described in section4.

We first define   $ X $  to represent the domain of the data samples,  $ \mathcal{F}^i $  to represent the $ i $-th sub-domain, $ \mathcal{F} $ to represent the domain of anomaly-free feature space. 

Let  $ f_{e_i}(.):X \rightarrow \mathcal{F}^i $  denote the i-th encoder and  $ f_d(.):  \mathcal{F} \rightarrow X $ denote the decoder.

Given a sample $ x \in X $, the encoder converts it to K encoded representations as  $ F_i \in \mathcal{F}^i $ ; and the decoder is trained to mapping anomaly-free latent representations  $ \hat F \in \mathcal{F} $   to the domain  $ X $   as follows
\begin{equation}\label{key}
F_i = f_{e_i}(x;\theta_{e_i}),i=1,2,...,K 
\end{equation}
\begin{equation}\label{key}
\hat{x} = f_d(\hat{F};\theta_d) 
\end{equation}

where $ \theta_{e_i} $  and  $ \theta_d $  denote the parameters of the $ i $-th encoder $ f_{e_i} $  and decoder $ f_d $  ,respectively. In the proposed multi-memAE, $ F_i $ is used to retrieve the relevant i-level memory items; and $ \hat F_i $  is the feature after memory addressing(anomaly-removal),  $ \hat F $ is the feature obtained by feature fusion. So we suppose taking the non-anomaly features  $ \hat F $ as input, the decoder will produce the de-anomaly image(see Fig.~\ref{fig:memae}).

In the training phase,we train our multi-memories AE on the normal data $ \{x^t\}_{t=1}^{T} $ , by conducting to minimize the reconstruction error on each sample: 
\begin{equation}\label{key}
R(x^t,\hat x^t) =||x^t-\hat x^t||_2^2 
\end{equation}
where the MSE-loss is used to updated the model parameters.
In the testing phase, we use the $ l_2 $-norm to measure the reconstruction quality at each pixel, then we use a error threshold $ e $ to decide which pixel may the anomalies locate at.

\subsection{Multi-Scale Memories}
\begin{figure}
	\centering
	\vspace{-0.8cm}
	\includegraphics[width=100mm]{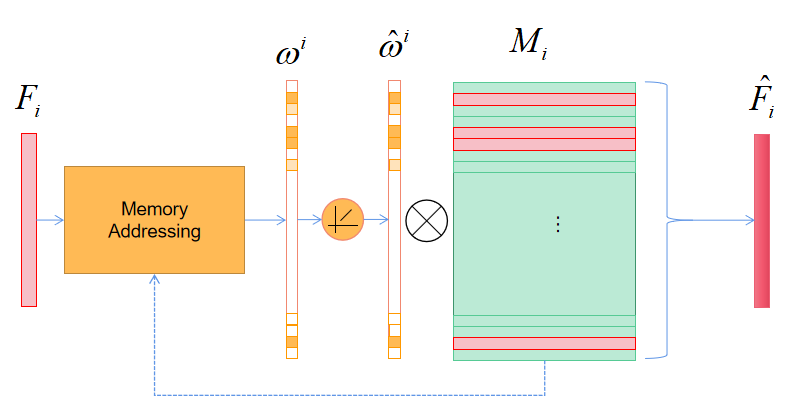} 
	\setlength{\abovecaptionskip}{-0.1cm}
	\caption{ The memory addressing unit takes the encoding $ F_i $ as query to obtain the soft addressing
		weights.After performing the hard shrinkage on weights, we got the weighted combination of slots in the memory $ M_i $.}
	\vspace{-0.5cm}
	\label{fig:memory_addressing}
\end{figure}
we adopt the memory addressing method in memAE(see Fig.~\ref{fig:memory_addressing}). According to memAE, slots in memory are updated to represent prototype features in normal data during training.Ideally, we want to replace our features with the slots in the memory that most closely resemble our features.
However, even in normal data space, there can be considerable variances between images, so we need to use combinations of slots to recover these differences.
Therefore, we compute the attention weights $ \omega $ based on the similarity of the memory items and the encoded feature.
Let's formalize the memory addressing process in  the proposed multi-memAE.
Taking the i-level feature  $ F_i $ and memory $ M_i $   as example,we first compute the similarity between  $ F_i $  and each slot   $ m_j^i\in M_i,j=1,....,n_i $as  
\begin{equation}
d(F_i,m_j^i) = \frac{F_i{m_j^i}^T}{\|F_i\| \|m_j^i\|} 
\end{equation}
then  $ \omega^i $  is obtained via a softmax operation:
\begin{equation}
\omega_j^i = \frac{exp(d(F_i,m^i_j))}{\sum_{t=1}^{n_i}exp(d(F_i,m^i_t))} 
\end{equation}
and the overall weights write as :
\begin{equation}
\omega = (\omega^1,...,\omega^K) 
\end{equation}
\begin{equation}
\omega^i = (\omega^i_1,...,\omega^i_{n_i}),i=1,...,K 
\end{equation}

We follow the tricks proposed by memAE--hard shrinkage and sparsity regularization to encourage the memory slots after updated to represent the prototype features.By performing hard shrinkage we got the final weights 
\begin{equation}
\hat\omega^i = (\hat\omega^i_1,...,\hat\omega^i_{n_i}),i=1,...,K 
\end{equation}

Finally ,we obtain  the de-anomalous feature  $ \hat F_i $  by combining the slots: 
\begin{equation}
\hat F_i = \sum_{j=1}^{n_i}\hat\omega^i_jm^i_j,i=1,...,K,j=1,...,n_i 
\end{equation}

\section{Experiments} 

\begin{table}[]
	\resizebox{\textwidth}{!}{
		\begin{tabular}{l|c|cccccccccc}
			\hline
			\multicolumn{2}{c|}{Category} & {\color[HTML]{000000} Ours} & {\color[HTML]{000000} 1-NN} & {\color[HTML]{000000} OC-SVM} & {\color[HTML]{000000} K-Means} & {\color[HTML]{000000} l2-AE} & {\color[HTML]{000000} VAE} & {\color[HTML]{000000} SSIM-AE} & {\color[HTML]{000000} MEM-AE} & {\color[HTML]{000000} AnoGAN} & {\color[HTML]{000000} \begin{tabular}[c]{@{}c@{}}GNN-Feature\\ Dictionary\end{tabular}} \\ \hline
			\multirow{5}{*}{\rotatebox{90}{Textures}}
			& Carpet     & \textbf{0.599} & 0.512 & 0.355  & 0.253   & 0.456          & 0.501 & 0.647          & 0.594          & 0.204  & 0.469  \\ \cline{2-12} 
			& Grid       & 0.610          & 0.228 & 0.125  & 0.107   & 0.582          & 0.224 & \textbf{0.849} & 0.519          & 0.226  & 0.183  \\ \cline{2-12} 
			& Leather    & \textbf{0.826} & 0.446 & 0.306  & 0.308   & 0.819          & 0.635 & 0.561          & 0.824          & 0.378  & 0.641  \\ \cline{2-12} 
			& Tile       & 0.582          & 0.822 & 0.722  & 0.779   & \textbf{0.897} & 0.87  & 0.175          & 0.588          & 0.177  & 0.797  \\ \cline{2-12} 
			& Wood       & 0.685          & 0.502 & 0.336  & 0.411   & \textbf{0.727} & 0.628 & 0.605          & 0.67           & 0.386  & 0.621  \\ \hline
			\multirow{10}{*}{\rotatebox{90}{Objectives}}
			& Bottle     & \textbf{0.916} & 0.898 & 0.85   & 0.495   & 0.910          & 0.897 & 0.87           & 0.74           & 0.62   & 0.742  \\ \cline{2-12} 
			& Cable      & 0.672          & 0.806 & 0.431  & 0.513   & \textbf{0.825} & 0.654 & 0.478          & 0.749          & 0.383  & 0.558  \\ \cline{2-12} 
			& Capsule    & \textbf{0.886} & 0.631 & 0.554  & 0.387   & 0.862          & 0.526 & 0.86           & 0.874          & 0.306  & 0.306  \\ \cline{2-12} 
			& Hazelnut   & \textbf{0.923} & 0.861 & 0.616  & 0.698   & 0.917          & 0.878 & 0.916          & 0.919          & 0.698  & 0.844  \\ \cline{2-12} 
			& Metal nut  & \textbf{0.869} & 0.705 & 0.319  & 0.351   & 0.83           & 0.576 & 0.603          & 0.807          & 0.32   & 0.358  \\ \cline{2-12} 
			& Pill       & \textbf{0.895} & 0.725 & 0.544  & 0.514   & 0.893          & 0.769 & 0.83           & 0.882          & 0.776  & 0.460  \\ \cline{2-12} 
			& Screw      & 0.865          & 0.604 & 0.644  & 0.55    & 0.754          & 0.559 & 0.887          & \textbf{0.896} & 0.466  & 0.277  \\ \cline{2-12} 
			& Toothbrush & \textbf{0.948} & 0.675 & 0.538  & 0.337   & 0.822          & 0.693 & 0.784          & 0.929          & 0.749  & 0.151  \\ \cline{2-12} 
			& Transistor & \textbf{0.751} & 0.68  & 0.496  & 0.399   & 0.728          & 0.626 & 0.725          & 0.746          & 0.549  & 0.628  \\ \cline{2-12} 
			& Zipper     & 0.677          & 0.512 & 0.355  & 0.253   & \textbf{0.839} & 0.549 & 0.665          & 0.684          & 0.467  & 0.703  \\ \hline
			\multicolumn{2}{c|}{Mean} 
			& 0.777          & 0.64  & 0.479  & 0.423   & \textbf{0.790} & 0.639 & 0.649          & 0.761          & 0.443  & 0.515                  \\ \hline
		\end{tabular}}

\caption{Results on the MVTec Anomaly Detection dataset. For each dataset category, the normalized area under the PRO-curve up to an average false positive rate per-pixel of 30$ \% $ is given. It measures the average overlap of each ground-truth region with the
	predicted anomaly regions for multiple thresholds. The best-performing method for each dataset category is highlighted in boldface.}
\label{Table:MVTec_result}
\end{table}

In this section, we validate the proposed MMAE for anomaly detection.To show the generality and applicability of the proposed model, we conduct experiments on both image and video datasets for anomaly detection. The results are compared with different baseline models and state-of-the-art techniques.The proposed MMAE is applied to all datasets following previous sections and we will elaborate our model specification in the following section. 
\subsection{Experiments on Image Anomaly Detection}

\begin{figure}
	\centering
	\includegraphics[width=120mm]{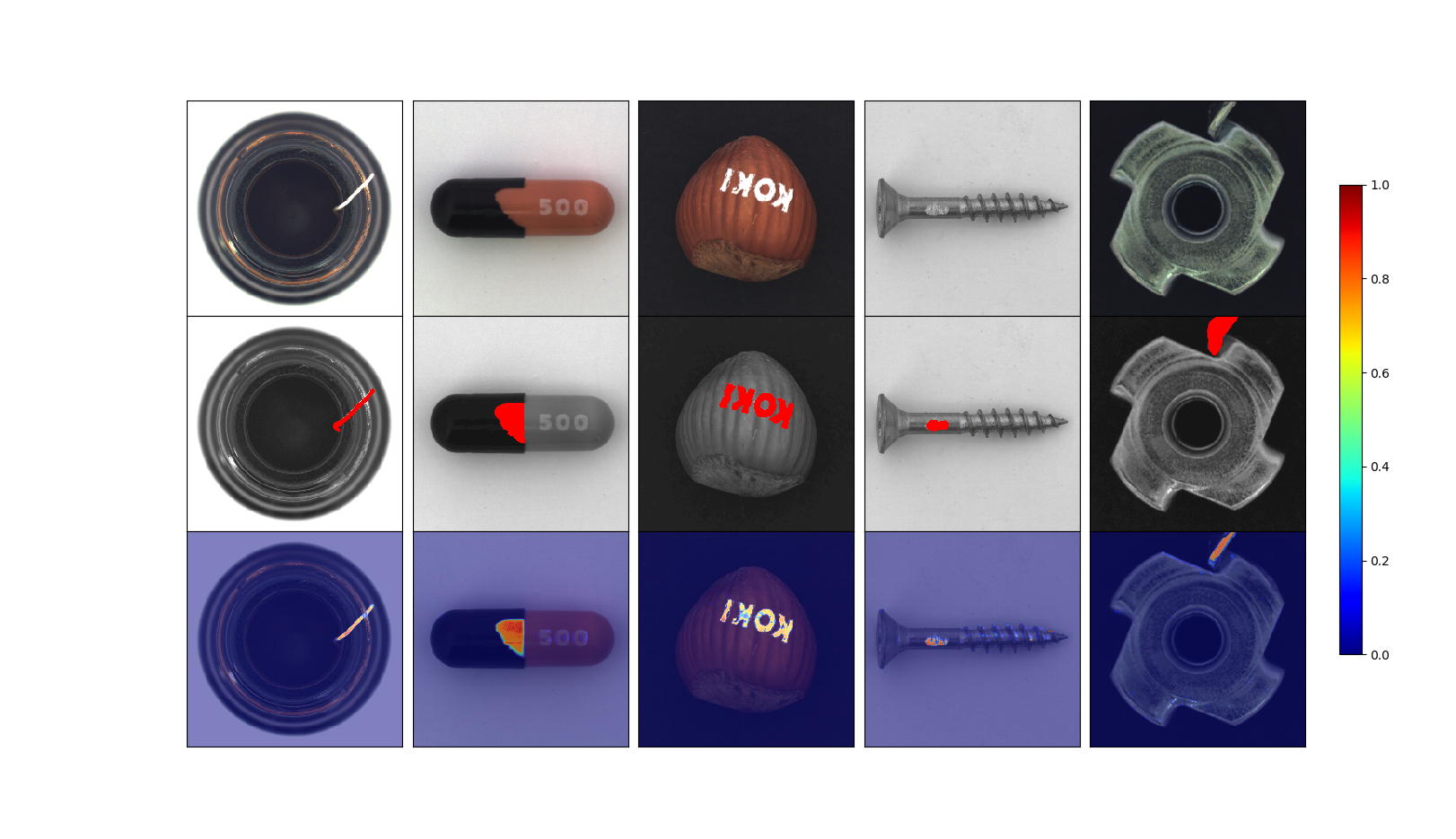} 
	\setlength{\abovecaptionskip}{-0.1cm}
	\caption{ Qualitative results of our anomaly detection method
		on the MVTec Anomaly Detection dataset. Top row: Defective
		input images. Center row: Ground truth regions of defects
		in red. Bottom row: Anomaly scores for each image pixel
		predicted by our algorithm.}
	\vspace{-0.5cm}
	\label{fig:detection results}
\end{figure}

We first conduct the experiments to pixel-precise anomaly segmentation and evaluate the performance on the newly published MVTec AD.

The MVTec Anomaly Detection (MVTec AD) dataset contains 5354 high-resolution color images of different object and texture categories.It contains normal images intended for training and images with anomalies intended for testing.The anomalies manifest themselves in the form of over 70 different types of defects such as scratches,dents, contaminations, and various structural changes. 

For all our experiments on MVTec AD, we set $ K = 5 $ and $ \gamma = 0.5 $, and input image scale zoomed to size $ w = h = 256 $ pixels.We train for $ 2000 $ epochs and batchsize $ 64 $, and the setting was same to all method used for comparison.We use the Adam optimizer at initial learning rate $10^{-4}$  and weight decay $10^{-5}$. 

For our experiments, We adopt strided $3 \times 3$ convolutions with the stride $ 2 $ for $2\times$ downsampling,and we use the pixel shuffle\cite{shi2016real} for $2\times$ upsampling.All architectures are simple CNNs with only convolutional and linear layers, using leaky rectifified linear units (LReLUs) with slope $ 0.2 $ as the activation function. The resolution levels are set as $ 16,32,64,128,256 $,memory sizes for MVTec AD are set as $ 60 $,the latent dim are set as $ 1024 $.The encoding module contains $ 5 $ encoder corresponding to each resolution and decoding module contains $ 1 $ decoder.

In addition to the memAE, we benchmark our approach against the best performing deep learning based methods presented by Bergmann et al. on this dataset. Specififically, these methods include the CNN-Feature Dictionary\cite{napoletano2018anomaly}, the SSIM-Autoencoder\cite{bergmann2018improving},and AnoGAN\cite{schlegl2017unsupervised}. 

We compute a threshold-independent measure based on the per-region-overlap (PRO) as the evaluation metric.It weights ground-truth regions of different size equally,which is in contrast to simple per-pixel measures for which a single large correctly segmented region can make up for many incorrectly segmented small ones. It was also used by Bergmann et al. in \cite{bergmann2019mvtec}.For computing the PRO metric, anomaly maps are first thresholded at a given anomaly score to make a binary decision for each pixel whether an anomaly is present or not(see Fig.~\ref{fig:detection results}). For each connected component within the ground-truth, the percentage of overlap with the thresholded anomaly region is computed.

We evaluate the PRO value for a large number of increasing thresholds until an average per pixel false positive rate of $30\%$ for the entire dataset is reached and integrate the area under the PRO curve as a measure of anomaly detection performance. Note that for high false positive rates, large parts of the input images would be wrongly labeled as anomalous and even perfect PRO values of 1.0 would no longer be meaningful. We normalize the integrated area to a maximum achievable value of 1.0.

Our method consistently outperforms other evaluated reconstruction-based algorithms for almost every object dataset category(see Table.~\ref{Table:MVTec_result}).However,we failed to get the best performance on texture datasets,we suppose that due to the representation space is constrained by the limited number of slots.

\begin{table}[]
	\centering
	\begin{tabular}{c|c|l|l|l|l|cllcllcll}
		
		\hline
		
		\multicolumn{6}{p{3cm}<{\centering}|}{Method\textbackslash{}Dataset} & \multicolumn{3}{p{2.5cm}<{\centering}|}{UCSD-Ped2} & \multicolumn{3}{p{2.5cm}<{\centering}|}{CUHK} & \multicolumn{3}{p{2.5cm}<{\centering}}{SH.Tech} \\ \hline
		
		\multirow{7}{*}{\rotatebox{90}{Non-Recon.}}
		& \multicolumn{5}{c|}{MPPCA}      & \multicolumn{3}{c}{0.693}          & \multicolumn{3}{c}{-}              & \multicolumn{3}{c}{-}              \\
		& \multicolumn{5}{c|}{MPPCA+SFA}  & \multicolumn{3}{c}{0.613}          & \multicolumn{3}{c}{-}              & \multicolumn{3}{c}{-}              \\
		& \multicolumn{5}{c|}{MDF}        & \multicolumn{3}{c}{0.829}          & \multicolumn{3}{c}{-}              & \multicolumn{3}{c}{-}              \\
		& \multicolumn{5}{c|}{AMDN}       & \multicolumn{3}{c}{0.908}          & \multicolumn{3}{c}{-}              & \multicolumn{3}{c}{-}              \\
		& \multicolumn{5}{c|}{Unmasking}  & \multicolumn{3}{c}{0.822}          & \multicolumn{3}{c}{0.806}          & \multicolumn{3}{c}{-}              \\
		& \multicolumn{5}{c|}{MT-FRCN}    & \multicolumn{3}{c}{0.922}          & \multicolumn{3}{c}{-}              & \multicolumn{3}{c}{-}              \\
		& \multicolumn{5}{c|}{Frame-Pred} & \multicolumn{3}{c}{\textbf{0.954}} & \multicolumn{3}{c}{\textbf{0.849}} & \multicolumn{3}{c}{\textbf{0.728}} \\ \hline
		
		\multirow{7}{*}{\rotatebox{90}{Recon.}}
		& \multicolumn{5}{c|}{AE-Conv2D}  & \multicolumn{3}{c}{0.850}          & \multicolumn{3}{c}{0.800}          & \multicolumn{3}{c}{0.609}          \\
		& \multicolumn{5}{c|}{AE-Conv3D}  & \multicolumn{3}{c}{0.912}          & \multicolumn{3}{c}{0.771}          & \multicolumn{3}{c}{-}              \\
		& \multicolumn{5}{c|}{TSC}        & \multicolumn{3}{c}{0.910}          & \multicolumn{3}{c}{0.806}          & \multicolumn{3}{c}{0.679}          \\
		& \multicolumn{5}{c|}{StackRNN}   & \multicolumn{3}{c}{0.922}          & \multicolumn{3}{c}{0.817}          & \multicolumn{3}{c}{0.680}          \\
		& \multicolumn{5}{c|}{AE}         & \multicolumn{3}{c}{0.917}          & \multicolumn{3}{c}{0.810}          & \multicolumn{3}{c}{0.697}          \\ 
		
		\cline{2-15} 
		& \multicolumn{5}{c|}{MemAE}      & \multicolumn{3}{c}{0.941}          & \multicolumn{3}{c}{0.833}          & \multicolumn{3}{c}{0.712}          \\
		& \multicolumn{5}{c|}{MMAE}       & \multicolumn{3}{c}{\textbf{0.949}} & \multicolumn{3}{c}{\textbf{0.846}} & \multicolumn{3}{c}{\textbf{0.731}} \\
		\hline
	\end{tabular}
\caption{AUC of different methods on video datasets UCSD-Ped2,
	CUHK Avenue and ShanghaiTech.}
\label{Table:video_detection_result}
\end{table}

\subsection{Experiments on Video Anomaly Detecion}
Video anomaly detection aims to identify abnormal content and movement patterns in video, which is an important task in video surveillance. We experiment on video anomaly detection datasets for three real-life scenarios, namely UCSD-Ped2\cite{mahadevan2010anomaly}, CUHK Avenue\cite{lu2013abnormal} and ShanghaiTech\cite{luo2017revisit}. Specifically, the latest benchmark dataset ShanghaiTech contains over 270,000 training frames and over 42, 000 frames (of which there are approximately 17,000 anomalous frames) for testing, which covered 13 different scenarios. In these datasets, objects other than pedestrians (e.g., vehicles) and strenuous movements (e.g., fights and chases) were treated as anomalies.
In order to preserve the spatio-temporal information of the video, we use 3D convolution to implement an encoder and decoder to extract the spatio-temporal features of the video. Thus, the input to the network is a cube of 16 grayscale adjacent frames stacked on top of each other.We set the K = 4.The single encoder,decoder,and memory module are setted following the configuration in memAE,except the number of channels after first convolution,which is designed to obtain the feature maps with same size.

Due to the complexity of the video data, many general anomaly detection methods\cite{parzen1962estimation,kingma2013auto,zong2018deep} that have not been specifically designed for this purpose cannot works well on video. To show the effectiveness of the proposed memory module, we compare the proposed MMAE with memAE and many well-designed reconstruction-based most Advanced methods are compared, including AE methods with 2D\cite{hasan2016learning} and 3D convolution \cite{zhao2017spatio} (AE-Conv2D and AE-Conv3D), a sparse coding method for spatiotemporal coherence (TST) \cite{luo2017revisit}, a stacked-loop Neural Network (StackRNN)\cite{luo2017revisit} and many video anomaly detection baselines.
Table.~\ref{Table:video_detection_result} shows the AUC values on video datasets.MMAE produces much better results than MemAE. The proposed MemAE obtains better or comparative performance than other methods.
\section{Conclusion}
Our method improves the performance on anomaly detection by introducing multi-scale memory modules and attention fuser while preserving the basic structure of MemAE, so that features at different scales of the data can be taken into account simultaneously.So it can be applied into end to end unsupervised anomaly detection requiring high precision.

\bibliographystyle{unsrt}  


\end{document}